\documentclass[10pt,twocolumn,letterpaper]{article}

\usepackage{cvpr}
\usepackage{times}
\usepackage{epsfig}
\usepackage{graphicx}
\usepackage{caption}
\usepackage{subcaption}
\usepackage{amsmath}
\usepackage{amssymb}
\usepackage{booktabs}
\usepackage{mathtools}
\usepackage{nicefrac}
\usepackage{algorithm,algpseudocode}
\usepackage{enumitem}

\usepackage[pagebackref=true,breaklinks=true,letterpaper=true,colorlinks,bookmarks=false]{hyperref}

\newcommand{\R}{\mathbb{R}}
\newcommand{\ra}[1]{\renewcommand{\arraystretch}{#1}}
\newcommand{\argmax}[1]{\mathop{\hbox{argmax}}_{#1}}
\newcommand{\argmin}[1]{\mathop{\hbox{argmin}}_{#1}}
\newcommand{\arginf}[1]{\mathop{\hbox{arginf}}_{#1}}

\newif\ifsupplementary
\supplementarytrue % Uncomment this line to include supplmentary section

\cvprfinalcopy % *** Uncomment this line for the final submission

\def\cvprPaperID{1079} % *** Enter the CVPR Paper ID here

\begin{document}

%%%%%%%%% TITLE
\title{Fast Patch-based Style Transfer of Arbitrary Style}

\author{Tian Qi Chen\\
Department of Computer Science\\
University of British Columbia\\
{\tt\small tqichen@cs.ubc.ca}
% For a paper whose authors are all at the same institution,
% omit the following lines up until the closing ``}''.
% Additional authors and addresses can be added with ``\and'',
% just like the second author.
% To save space, use either the email address or home page, not both
\and
Mark Schmidt\\
Department of Computer Science\\
University of British Columbia\\
{\tt\small schmidtm@cs.ubc.ca}
}

\maketitle
%\thispagestyle{empty}

%%%%%%%%% ABSTRACT
\begin{abstract}
Artistic style transfer is an image synthesis problem where the content of an image is reproduced with the style of another. Recent works show that a visually appealing style transfer can be achieved by using the hidden activations of a pretrained convolutional neural network. However, existing methods either apply (i) an optimization procedure that works for any style image but is very expensive, or (ii) an efficient feedforward network that only allows a limited number of trained styles. In this work we propose a simpler optimization objective based on local matching that combines the content structure and style textures in a single layer of the pretrained network. We show that our objective has desirable properties such as a simpler optimization landscape, intuitive parameter tuning, and consistent frame-by-frame performance on video. Furthermore, we use 80,000 natural images and 80,000 paintings to train an inverse network that approximates the result of the optimization. This results in a procedure for artistic style transfer that is efficient but also allows arbitrary content and style images.
\end{abstract}

\section{Introduction}
%\begin{figure}
%	\centering
%	\begin{subfigure}[b]{0.49\linewidth}
%		\centering
%		\includegraphics[width=\linewidth]{images/boat_orig_crop.jpg}
%		\caption*{Content}
%	\end{subfigure}%
%	\hfill
%	\begin{subfigure}[b]{0.49\linewidth}
%		\centering
%		\includegraphics[height=33mm]{images/small_worlds_one.jpg}
%		\caption*{Style}
%	\end{subfigure}\\
%	\begin{subfigure}[b]{0.49\linewidth}
%		\centering
%		\includegraphics[width=\linewidth]{images/boat_stylized_crop.jpg}
%		\caption*{Our Style Transfer}
%	\end{subfigure}%
%	\hfill
%	\begin{subfigure}[b]{0.49\linewidth}
%		\centering
%		\includegraphics[width=\linewidth]{images/boat_stylized_dec_crop.jpg}
%		\caption*{Our Feed-Forward Approx.}
%	\end{subfigure}
%	\caption{An example of our artistic style transfer method and its feedforward approximation. The approximation network has never seen this style image during training.}
%	\label{fig:showcase}
%\end{figure}

\begin{figure}
	\centering
	\begin{subfigure}[b]{0.49\linewidth}
		\centering
		\includegraphics[width=\linewidth]{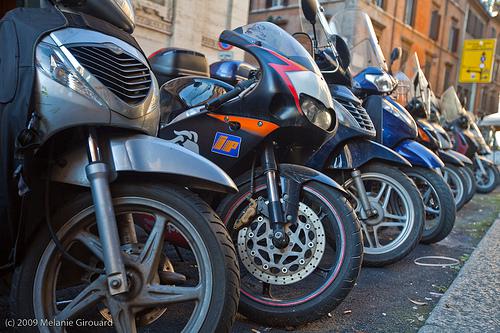}
		\caption*{Content}
	\end{subfigure}%
	\hfill
	\begin{subfigure}[b]{0.49\linewidth}
		\centering
		\includegraphics[width=\linewidth]{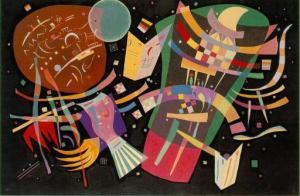}
		\caption*{Style}
	\end{subfigure}\\
	\begin{subfigure}[b]{0.49\linewidth}
		\centering
		\includegraphics[width=\linewidth]{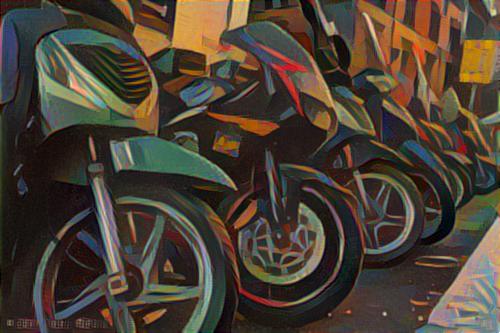}
		\caption*{Our Style Transfer}
	\end{subfigure}%
	\hfill
	\begin{subfigure}[b]{0.49\linewidth}
		\centering
		\includegraphics[width=\linewidth]{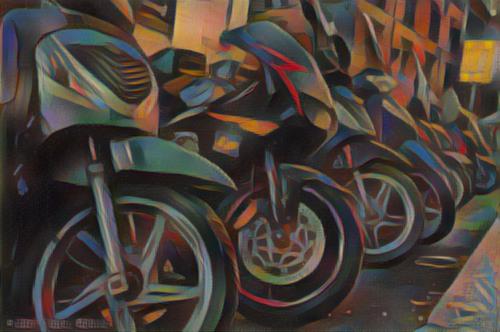}
		\caption*{Our Feed-Forward Approx.}
	\end{subfigure}
	\caption{An example of our artistic style transfer method and its feedforward approximation. The approximation network has never seen the content or style image during training.}
	\label{fig:showcase}
\end{figure}

Famous artists are typically renowned for a particular artistic style, which takes years to develop. Even once perfected, a single piece of art can take days or even months to create. This motivates us to explore efficient computational strategies for creating  artistic images. While there is a large classical literature on texture synthesis methods that create artwork from a blank canvas~\cite{efros1999texture,kwatra2005texture,liang2001real,wei2000fast}, several recent approaches study the problem of transferring the desired \emph{style} from one image onto the structural \emph{content} of another image. This approach is known as artistic style transfer.

The vague notion of artistic style is difficult to quantitatively capture. Early works define style using similarity measures or local statistics based on the pixel values~\cite{efros2001image,efros1999texture,Hertzmann2001,hertzmann2001image,kwatra2005texture,Litwinowicz1997}. However, recent methods that achieve impressive visual quality come from the use of convolutional neural networks (CNN) for feature extraction~\cite{elad2016style,frigo2016split,GatysEB15a,Li2016}. The success of these methods has even created a market for mobile applications that can stylize user-provided images on demand~\cite{artify,prisma,picsart}.

%Pixel representations with three channels only capture information for the color at a single pixel. On the other hand, deep representations in a CNN describe an image using hundreds of channels, where the values at each spatial location can also represent local statistics that describe the textures around the pixel. The power of this deep representation may be due to the architecture of the CNN~\cite{HeWH16}, with stacked filters that convolve the image multiple times. 

Despite this renewed interest, the actual process of style transfer is based on solving a complex optimization procedure, which can take minutes on today's hardware. A typical speedup solution is to train another neural network that approximates the optimum of the optimization in a single feed-forward pass~\cite{DumoulinSK16,Johnson2016,Ulyanov2016,DBLP:UlyanovVL16}. While much faster, existing works that use this approach sacrifice the versatility of being able to perform style transfer with any given style image, as the feed-forward network cannot generalize beyond its trained set of images. Due to this limitation, existing applications are either time-consuming or limited in the number of provided styles, depending on the method of style transfer.

In this work we propose a method that addresses these limitations: a new method for artistic style transfer that is efficient but is not limited to a finite set of styles. To accomplish this, we define a new optimization objective for style transfer that notably only depends on one layer of the CNN (as opposed to existing methods that use multiple layers). The new objective  leads to visually-appealing results while this simple restriction allows us to use an ``inverse network" to deterministically invert the activations from the stylized layer to yield the stylized image.

Section~\ref{sec:relatedwork} reviews related work while Sections~\ref{sec:method}-\ref{sec:inverse} describe our  new optimization objective that combines style and content statistics in a single activation layer. We then propose a method of training a neural network that can invert the activations to yield an image. Section~\ref{sec:experiments} presents experiments with the new method, showing it has desirable properties not found in existing formulations. %Additionally, we show that our trained network does produce an approximation to the optimal image. In Section~\ref{sec:discussion}, we conclude our paper and propose possible future research directions.

\section{Related Work}\label{sec:relatedwork}
%We describe existing methods that make use of a convolutional neural network for feature extraction. Note that each proposed work has a specific goal, such as arbitrary-style versatility, application to video, or speed improvement. However, no existing work contains all three properties.

\textbf{Style Transfer as Optimization.} Gatys \etal \cite{GatysEB15a} formulates style transfer as an optimization problem that combines texture synthesis with content reconstruction. Their formulation involves additive loss functions placed on multiple layers of a pretrained CNN, with some loss functions synthesizing the textures of the style image and some loss functions reconstructing the content image. Gradients are computed by backpropagation and gradient-based optimization is  carried out to solve for the stylized image. An alternative approach uses patch-based similarity matching between the content and style images \cite{elad2016style,frigo2016split,Li2016}. In particular, Li and Wand \cite{Li2016} constructs patch-based loss functions, where each synthetic patch has a nearest neighbour target patch that it must match. This type of patch-matching loss function is then combined additively with Gatys \etal's formulation. While these approaches allow arbitrary style images, the optimization frameworks used by these methods makes it slow to generate the stylized image. This is particularly relevant for the case of video where we want to style a huge number of frames.

\textbf{Feed-forward Style Networks} As mentioned previously, it is possible to train a neural network that approximates the optimum of Gatys \etal's loss function for one or more fixed styles~\cite{DumoulinSK16,Johnson2016,Ulyanov2016,DBLP:UlyanovVL16}. 
%These approaches specifically optimize Gatys \etal's loss function. 
This yields a much faster method, but these methods need to be re-trained for each new style.

\textbf{Style Transfer for Video.} Ruder \etal \cite{Ruder2016} introduces a temporal loss function that, when used additively with Gatys \etal's loss functions, can perform style transfer for videos with temporal consistency. Their loss function relies on optical flow algorithms for gluing the style in place across nearby frames. This results in an order of maginitude slowdown compared to frame-by-frame application of style transfer.

\textbf{Inverting Deep Representations.} Several works have trained inverse networks of pretrained convolutional neural networks \cite{DosovitskiyB15,mahendran2015understanding} for the goal of visualization. Other works have trained inverse networks as part of an autoencoder \cite{kulkarni2015deep,masci2011stacked,zeiler2010deconvolutional}. To the best of our knowledge, all existing inverse networks are trained with a dataset of images and a loss function placed in RGB space.

In comparison to existing style transfer approaches, we propose a method for directly constructing the target activations for a single layer in a pretrained CNN. Like Li and Wand \cite{Li2016}, we make use of a criteria for finding best matching patches in the activation space. However, our method is able to directly construct the entire activation target. This deterministic procedure allows us to easily adapt to video, without the need to rely on optical flow for fixing consistency issues. In addition to optimization, we propose a feed-forward style transfer procedure by inverting the pretrained CNN. Unlike existing feed-forward style transfer approaches, our method is not limited to specifically trained styles and can easily adapt to arbitary content and style images. Unlike existing CNN inversion methods, our method of training does not use a pixel-level loss, but instead uses a loss on the activations. By using a particular training setup~(see Section~\ref{sec:augment}), this inverse network is even able to invert activations that are outside the usual range of the CNN activiations.

\section{A New Objective for Style Transfer}\label{sec:method}
The main component of our style transfer method is a patch-based operation for constructing the target activations in a single layer, given the style and content images. We refer to this procedure as ``swapping the style'' of an image, as the content image is replaced patch-by-patch by the style image. We first present this operation at a high level, followed by more details on our implementation.% a more detailed explanation specific to a method of implementation.

\subsection{Style Swap}\label{sec:styleswap}
Let $C$ and $S$ denote the RGB representations of the content and style images (respectively), and let $\Phi(\cdot)$ be the function represented by a fully convolutional part of a pretrained CNN that maps an image from RGB to some intermediate activation space. After computing the activations, $\Phi(C)$ and $\Phi(S)$, the \emph{style swap} procedure is as follows:
\begin{enumerate}
	\item Extract a set of patches for both content and style activations, denoted by $\{\phi_i(C)\}_{i\in n_c}$ and $\{\phi_j(S)\}_{j\in n_s}$, where $n_c$ and $n_s$ are the number of extracted patches. The extracted patches should have sufficient overlap, and contain all channels of the activations.
	\item For each content activation patch, determine a closest-matching style patch based on the normalized cross-correlation measure, 
	\begin{equation}
	\phi_i^{ss}(C, S) := \argmax{\phi_j(S), \,j=1,\dots,n_s} \frac{\langle \phi_i(C), \phi_j(S) \rangle}{||\phi_i(C)|| \cdot ||\phi_j(S)||}.
	\label{eq:bestmatch}
	\end{equation}
	\item Swap each content activation patch $\phi_i(C)$ with its closest-matching style patch $\phi_i^{ss}(C, S)$.
	\item Reconstruct the complete content activations, which we denote by $\Phi^{ss}(C, S)$, by averaging overlapping areas that may have different values due to step 3.
\end{enumerate}
This operation results in hidden activations corresponding to a single image with the structure of the content image, but with textures taken from the style image.

\subsection{Parallelizable Implementation}\label{sec:styleswap-impl}
\begin{figure}
	\centering
	\includegraphics[width=\linewidth]{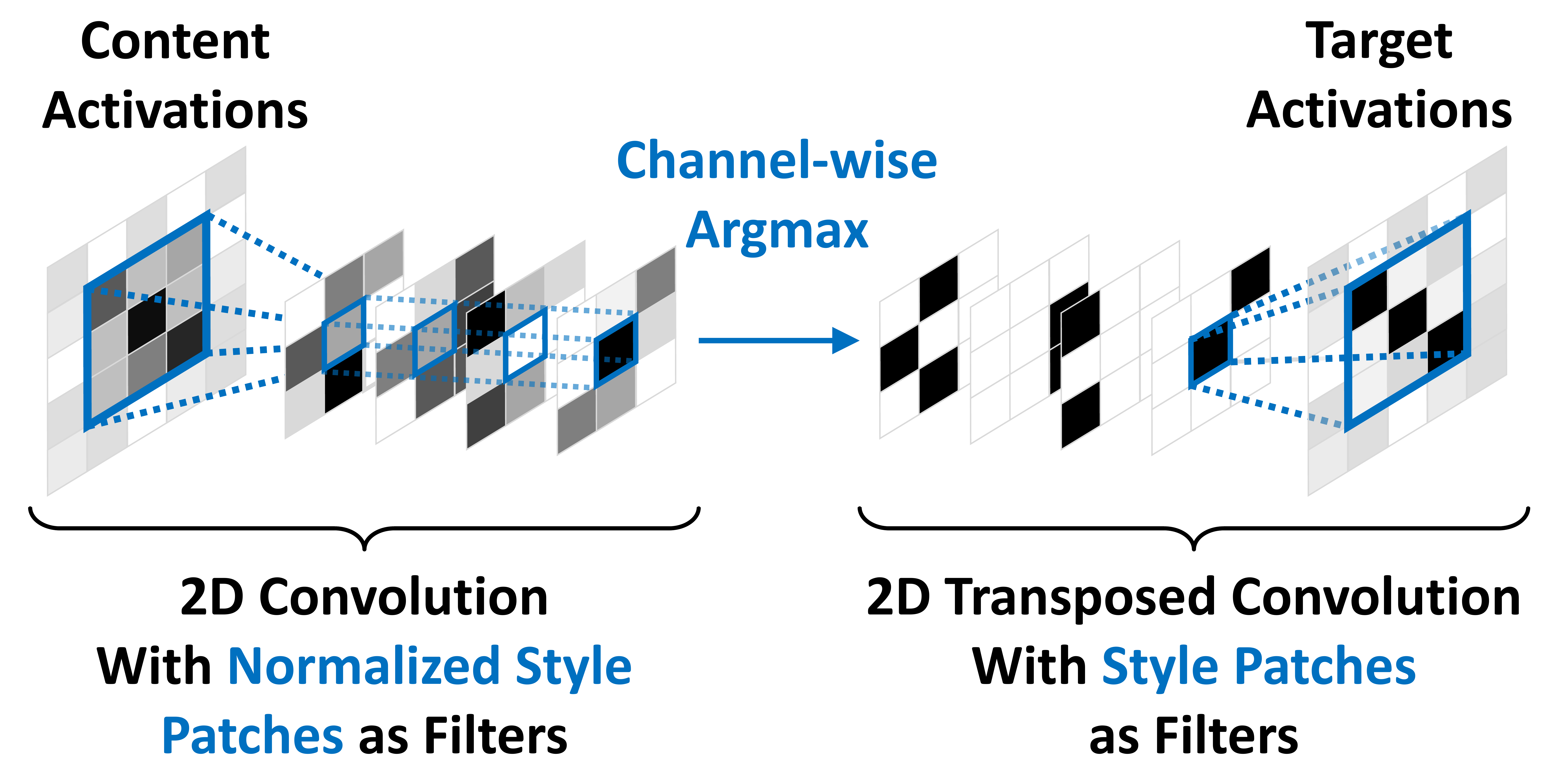}
	\caption{Illustration of a style swap operation. The 2D convolution extracts patches of size $3\times 3$ and stride $1$, and computes the normalized cross-correlations. There are $n_c=9$ spatial locations and $n_s=4$ feature channels immediately before and after the channel-wise argmax operation. The 2D transposed convolution reconstructs the complete activations by placing each best matching style patch at the corresponding spatial location.}
	\label{fig:styleswap} 
\end{figure}
To give an efficient implementation, we show that the entire style swap operation can be implemented as a network with three operations: (i) a 2D convolutional layer, (ii) a channel-wise argmax, and (iii) a 2D transposed convolutional layer. Implementation of style swap is then as simple as using existing efficient implementations of 2D convolutions and transposed convolutions\footnote{The transposed convolution is also often referred to as a ``fractionally-strided" convolution, a ``backward" convolution, an ``upconvolution", or a "deconvolution".}.

To concisely describe the implementation, we re-index the content activation patches to explicitly denote spatial structure. In particular, we'll let $d$ be the number of feature channels of $\Phi(C)$, and let $\phi_{a,b}(C)$ denote the patch $\Phi(C)_{a:a+s,\;b:b+s,\;1:d}$ where $s$ is the patch size.

Notice that the normalization term for content activation patches $\phi_i(C)$ is constant with respect to the argmax operation, so~\eqref{eq:bestmatch} can be rewriten as
\begin{equation}
\begin{split}
K_{a,b,j} =  \left\langle \phi_{a,b}(C), \frac{\phi_j(S)}{||\phi_j(S)||} \right\rangle \\
\phi_{a,b}^{ss}(C, S) = \argmax{\phi_j(S), \,j\in N_s} \{K_{a,b,j} \}
\end{split}
\label{eq:bestmatchsimple}
\end{equation}
The lack of a normalization for the content activation patches simplifies computation and allows our use of 2D convolutional layers. The following three steps describe our implementation and are illustrated in Figure~\ref{fig:styleswap}:
\begin{itemize}
	\item The tensor $K$ can be computed by a single 2D convolution by using the normalized style activations patches $\left\{\phi_j(S) / ||\phi_j(S)||\right\}$ as convolution filters and $\Phi(C)$ as input. The computed $K$ has $n_c$ spatial locations and $n_s$ feature channels. At each spatial location, $K_{a,b}$ is a vector of cross-correlations between a content activation patch and all style activation patches. 

	\item To prepare for the 2D transposed convolution, we replace each vector $K_{a,b}$ by a one-hot vector corresponding to the best matching style activation patch.
\begin{equation}
\overline{K}_{a,b,j} = \begin{cases}
1 & \quad \text{if $j= \argmax{j'}\{K_{a,b,j'}\}$} \\
0 & \quad \text{otherwise}
\end{cases}
\label{eq:argmax}
\end{equation}
	\item The last operation for constructing $\Phi^{ss}(C,S)$ is a 2D transposed convolution with $\overline{K}$ as input and unnormalized style activation patches $\left\{\phi_j(S)\right\}$ as filters. At each spatial location, only the best matching style activation patch is in the output, as the other patches are multiplied by zero.
\end{itemize}
Note that a transposed convolution will sum up the values from overlapping patches. In order to average these values, we perform an element-wise division on each spatial location of the output by the number of overlapping patches. Consequently, we do not need to impose that the argmax in~\eqref{eq:argmax} has a unique solution, as multiple argmax solutions can simply be interpreted as adding more overlapping patches. 
\begin{figure*}
\centering
\begin{subfigure}[b]{0.24\linewidth}
\centering
\includegraphics[width=\linewidth]{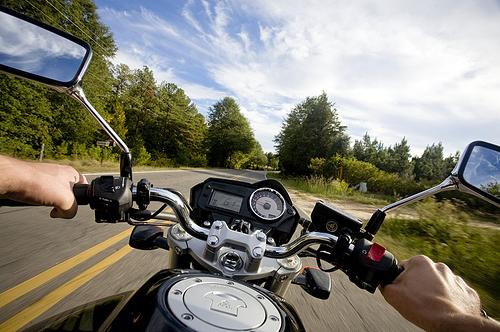}
\caption*{Content Image}
\end{subfigure}%
\hfill
\begin{subfigure}[b]{0.24\linewidth}
\centering
\includegraphics[width=\linewidth]{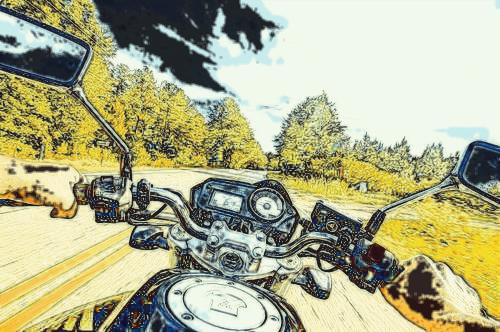}
\caption*{RGB}
\end{subfigure}%
\hspace{1pt}
\begin{subfigure}[b]{0.24\linewidth}
\centering
\includegraphics[width=\linewidth]{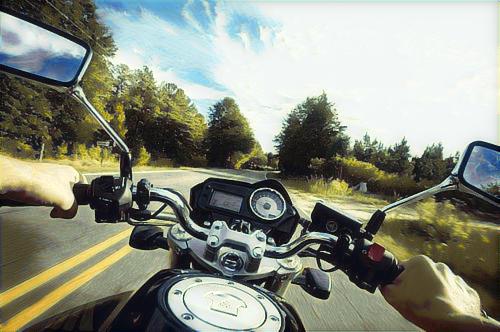}
\caption*{relu1\_1}
\end{subfigure}%
\hspace{1pt}
\begin{subfigure}[b]{0.24\linewidth}
\centering
\includegraphics[width=\linewidth]{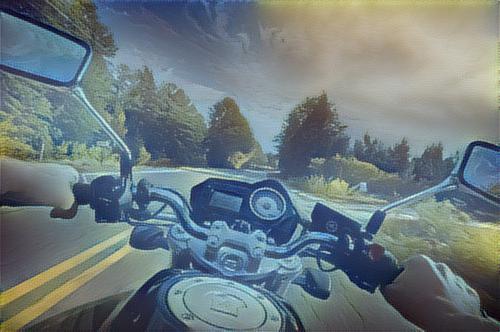}
\caption*{relu2\_1}
\end{subfigure}\\
\begin{subfigure}[b]{0.24\linewidth}
\centering
\includegraphics[width=\linewidth]{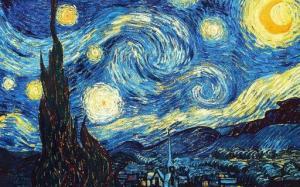}
\caption*{Style Image}
\end{subfigure}%
\hfill
\begin{subfigure}[b]{0.24\linewidth}
\centering
\includegraphics[width=\linewidth]{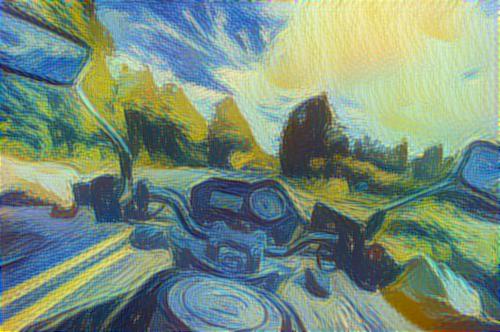}
\caption*{relu3\_1}
\end{subfigure}%
\hspace{1pt}
\begin{subfigure}[b]{0.24\linewidth}
\centering
\includegraphics[width=\linewidth]{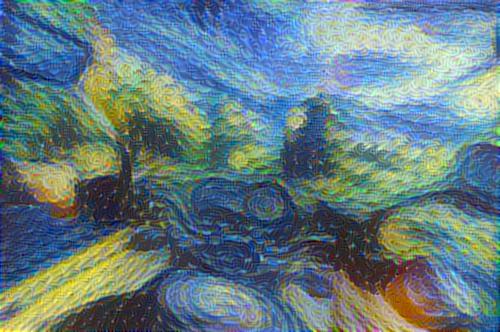}
\caption*{relu4\_1}
\end{subfigure}%
\hspace{1pt}
\begin{subfigure}[b]{0.24\linewidth}
\centering
\includegraphics[width=\linewidth]{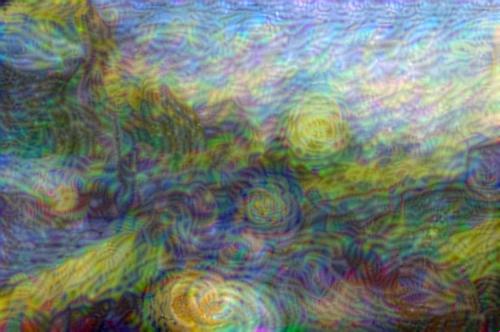}
\caption*{relu5\_1}
\end{subfigure}%
\caption{The effect of style swapping in different layers of VGG-19 \cite{simonyan2014very}, and also in RGB space. Due to the naming convention of VGG-19, ``relu$X$\_1'' refers to the first ReLU layer after the $(X-1)$-th maxpooling layer. The style swap operation uses patches of size $3\times 3$ and stride $1$, and then the RGB image is constructed using optimization.}
\label{fig:swaplayers}
\end{figure*}
\subsection{Optimization Formulation}\label{sec:optim}
The pixel representation of the stylized image can be computed by placing a loss function on the activation space with target activations $\Phi^{ss}(C,S)$. Similar to prior works on style transfer \cite{GatysEB15a,Li2016}, we use the squared-error loss and define our optimization objective as
\begin{equation}
\begin{split}
I_{stylized}(C, S) = \argmin{I\in\R^{h\times w\times d}} ||\Phi(I) - \Phi^{ss}(C, S)||_F^2 \\ + \lambda\ell_{TV}(I)
\end{split}
\label{eq:optloss}
\end{equation}
where we'll say that the synthesized image is of dimension $h$ by $w$ by $d$, $||\cdot||_F$ is the Frobenius norm, and $\ell_{TV}(\cdot)$ is the total variation regularization term widely used in image generation methods \cite{aly2005image,Johnson2016,mahendran2015understanding}. Because $\Phi(\cdot)$ contains multiple maxpooling operations that downsample the image, we use this regularization as a natural image prior, obtaining spatially smoother results for the re-upsampled image. The total variation regularization is as follows:
\begin{equation}
\begin{split}
\ell_{TV}(I) = \sum_{i=1}^{h-1} \sum_{j=1}^{w} \sum_{k=1}^d (I_{i+1,j,k} - I_{i,j,k})^2 \\ + \sum_{i=1}^{h} \sum_{j=1}^{w-1} \sum_{k=1}^d (I_{i,j+1,k} - I_{i,j,k})^2
\end{split}
\label{eq:tvloss}
\end{equation}
Since the function $\Phi(\cdot)$ is part of a pretrained CNN and is at least once subdifferentiable, \eqref{eq:optloss} can be computed using standard subgradient-based optimization methods. 

\section{Inverse Network}\label{sec:inverse}
Unfortunately, the cost of solving the optimization problem to compute the stylized image might be too high in applications such as video stylization. We can improve optimization speed by approximating the optimum using another neural network. Once trained, this network can then be used to produce stylized images much faster, and we will in particular train this network to have the versatility of being able to use new content and new style images.

The main purpose of our inverse network is to approximate an optimum of the loss function in~\eqref{eq:optloss} for any target activations. We therefore define the optimal inverse function as:
\begin{equation}
\begin{split}
\arginf{f}\; \mathbb{E}_{H} \Bigg[ ||\Phi(f(H)) - H||_F^2
+ \lambda \ell_{TV}\left(f(H)\right) \Bigg]
\end{split}
\label{eq:optimInvNet}
\end{equation}
where $f$ represents a deterministic function and $H$ is a random variable representing target activations. The total variation regularization term is added as a natural image prior similar to~\eqref{eq:optloss}. 

\subsection{Training the Inverse Network}\label{sec:augment}
A couple problems arise due to the properties of the pretrained convolutional neural network.

\textbf{Non-injective.} The CNN defining $\Phi(\cdot)$ contains convolutional, maxpooling, and ReLU layers. These functions are many-to-one, and thus do not have well-defined inverse functions. Akin to existing works that use inverse networks \cite{dosovitskiy2016learning,long2015fully,zeiler2010deconvolutional}, we instead train an approximation to the inverse relation by a parametric neural network. 
\begin{equation}
\begin{split}
\min_{\theta} \frac{1}{n} \sum_{i=1}^{n} ||\Phi(f(H_i; \theta)) - H_i||_F^2 
+ \lambda \ell_{TV}\left(f(H_i; \theta)\right)
\end{split}
\label{eq:trainInvNet}
\end{equation}
where $\theta$ denotes the parameters of the neural network $f$ and $H_i$ are activation features from a dataset of size $n$. This objective function leads to unsupervised training of the neural network as the optimum of \eqref{eq:optloss} does not need to be known. We place the description of our inverse network architecture in the appendix.

\textbf{Non-surjective.} The style swap operation produces target activations that may be outside the range of $\Phi(\cdot)$ due to the interpolation. This would mean that if the inverse network is only trained with real images then the inverse network may only be able to invert activations in the range of $\Phi(\cdot)$. Since we would like the inverse network to invert style swapped activations, we augment the training set to include these activations. More precisely, given a set of training images (and their corresponding activations), we augment this training set with style-swapped activations based on pairs of images. 

%The style swap operation produces target activations that are interpolated between output values of $\Phi(\cdot)$. A case may occur where the interpolated values are outside the range of $\Phi(\cdot)$. With regards to training an inverse network, this property would suggest that if we were to only train using real images---and obtain $H_i$ in~\eqref{eq:trainInvNet} by mapping real images to activation features---then at test time the network may be inverting activations that are the result of style swapping and thus be outside its trained domain.

%To ensure the network can invert style-swapped activations, we augment the training set to include these activations. More precisely, given a set of training images (and corresponding activations), we augment this training set with style-swapped activations based on pairs of images. This augmented set of activations is then used to train the inverse network.

\begin{figure}
	\centering
	\includegraphics[width=0.9\linewidth]{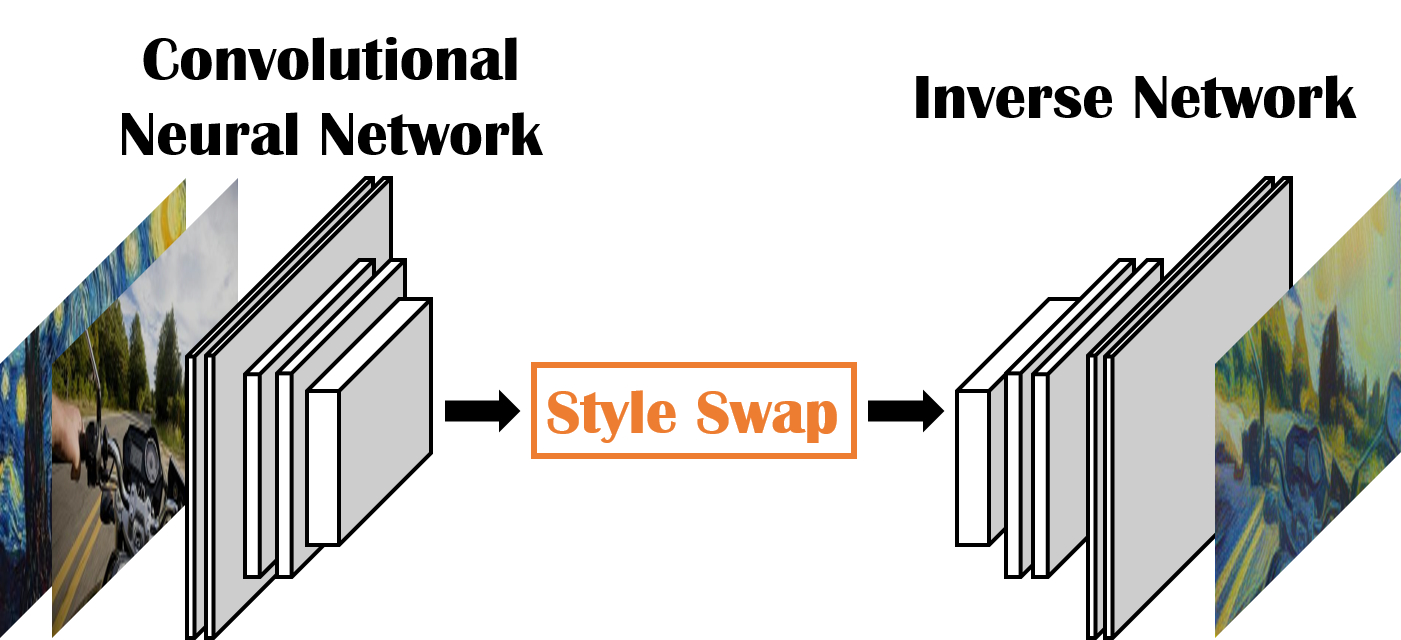}
	\caption{We propose the first feedforward method for style transfer that can be used for arbitrary style images. We formulate style transfer using a constructive procedure (Style Swap) and train an inverse network to generate the image.}
	\label{fig:feedforward-procedure}
\end{figure}

\subsection{Feedforward Style Transfer Procedure}\label{sec:ffstyle}
Once trained, the inverse network can be used to replace the optimization procedure. Thus our proposed feedforward procedure consists of the following steps:
\begin{enumerate}\vspace{-0mm}
	\item Compute $\Phi(C)$ and $\Phi(S)$.\vspace{-2mm}
	\item Obtain $\Phi^{ss}(C,S)$ by style swapping.\vspace{-2mm}
	\item Feed $\Phi^{ss}(C,S)$ into a trained inverse network.\vspace{-0mm}
\end{enumerate}
This procedure is illustrated in Figure~\ref{fig:feedforward-procedure}. As described in Section~\ref{sec:styleswap-impl}, style swapping can be implemented as a (non-differentiable) convolutional neural network. As such, the entire feedforward procedure can be seen as a neural net with individually trained parts. Compared to existing feedforward approaches \cite{DumoulinSK16,Johnson2016,Ulyanov2016,DBLP:UlyanovVL16}, the biggest advantage of our feedforward procedure is the ability to use new style images with only a single trained inverse network.

\section{Experiments}\label{sec:experiments}
In this section, we analyze properties of the proposed style transfer and inversion methods. We use the Torch7 framework \cite{torch} to implement our method\footnote{Code available at https://github.com/rtqichen/style-swap}, and use existing open source implementations of prior works \cite{Johnson2015,Li2016,Ruder2016} for comparison.

\subsection{Style Swap Results}

\begin{figure}
\centering
\begin{subfigure}[b]{0.3\linewidth}
\centering
\includegraphics[width=\linewidth]{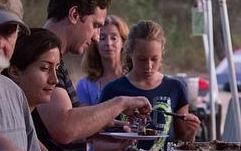}
\caption*{Content Image}
\end{subfigure}%
\hfill
\begin{subfigure}[b]{0.69\linewidth}
\centering
\includegraphics[width=0.49\linewidth]{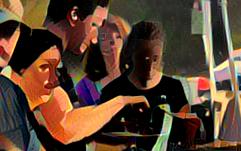}
\includegraphics[width=0.49\linewidth]{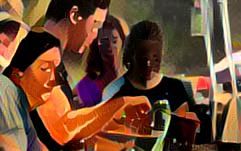}
\caption*{Gatys \etal with random initializations}
\end{subfigure}\\
\begin{subfigure}[b]{0.3\linewidth}
\centering
\includegraphics[width=\linewidth]{images/wassily_kandinsky.jpg}
\caption*{Style Image}
\end{subfigure}%
\hfill
\begin{subfigure}[b]{0.69\linewidth}
\centering
\includegraphics[width=0.49\linewidth]{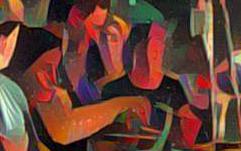}
\includegraphics[width=0.49\linewidth]{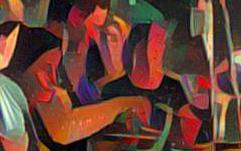}
\caption*{Our method with random initializations}
\end{subfigure}\\
\caption{Our method achieves consistent results compared to existing optimization formulations. We see that Gatys \etal's formulation \cite{GatysEB15a} has multiple local optima while we are able to consistently achieve the same style transfer effect with random initializations. Figure~\ref{fig:quantconsistency} shows this quantitatively.}
\label{fig:consistency}
\end{figure}

\begin{figure}
\centering
\includegraphics[width=0.95\linewidth,trim={90px 270px 110px 270px},clip]{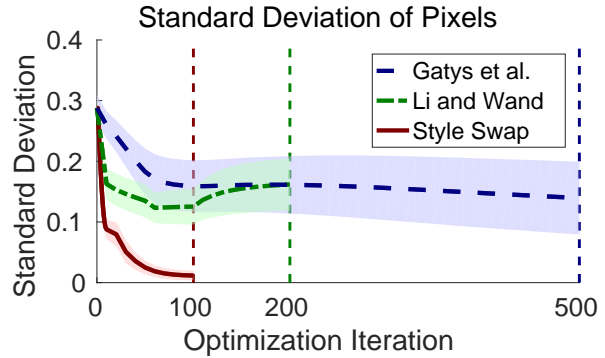}
\caption{Standard deviation of the RGB pixels over the course of optimization is shown for 40 random initializations. The lines show the mean value and the shaded regions are within one standard deviation of the mean. The vertical dashed lines indicate the end of optimization. Figure~\ref{fig:consistency} shows examples of optimization results.}
\label{fig:quantconsistency}
\end{figure}

\begin{figure*}
	\centering
	
	\begin{subfigure}[b]{0.5\linewidth}
		
		\hspace{10mm}
		\raisebox{7mm}{
			\begin{minipage}{30mm}
				\centering
				\textbf{Style} \\
				\textit{Small Worlds I}, Wassily Kandinsky, 1922
			\end{minipage}
		}
		\includegraphics[height=17mm]{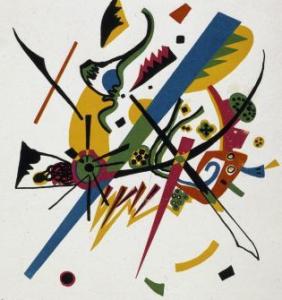}
	\end{subfigure}%
	\begin{subfigure}[b]{0.5\linewidth}
		\hspace{4mm}
		\raisebox{7mm}{
			\begin{minipage}{30mm}
				\centering
				\textbf{Style} \\
				\textit{The Starry Night}, Vincent Van Gogh, 1889
			\end{minipage}
		}
		\includegraphics[height=17mm]{images/starry_night.jpg}
	\end{subfigure}\\
	\begin{subfigure}[b]{1\linewidth}
		\centering
		\includegraphics[width=0.15\linewidth]{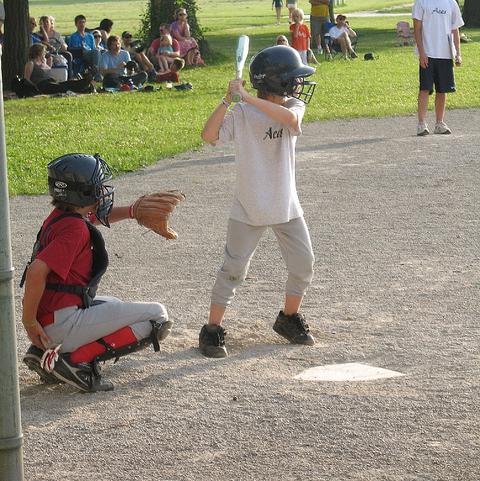}
		\includegraphics[width=0.15\linewidth]{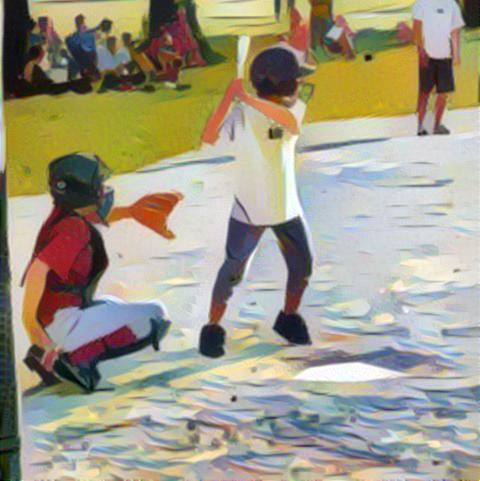}
		\includegraphics[width=0.15\linewidth]{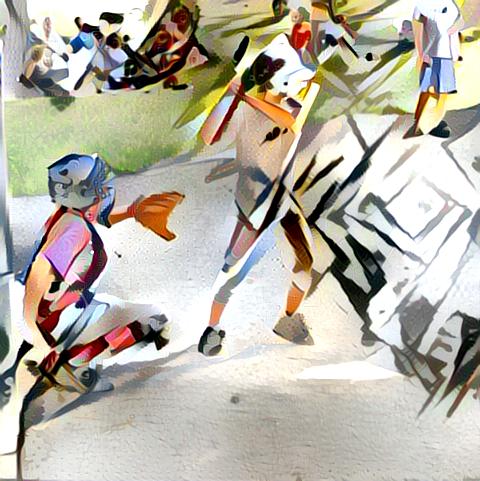}
		\includegraphics[width=0.15\linewidth]{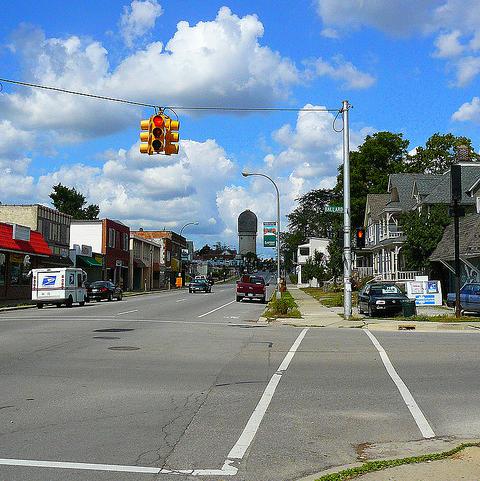}
		\includegraphics[width=0.15\linewidth]{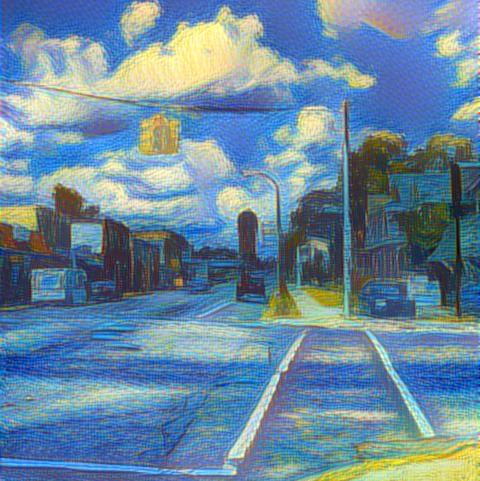}
		\includegraphics[width=0.15\linewidth]{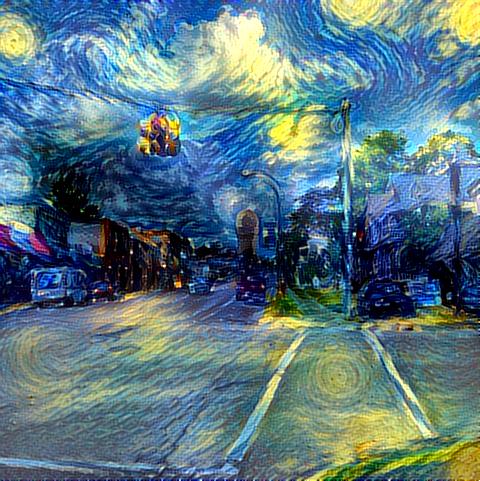}
	\end{subfigure} \\
	\vspace{1px}
	\begin{subfigure}[b]{1\linewidth}
		\centering
		\includegraphics[width=0.15\linewidth]{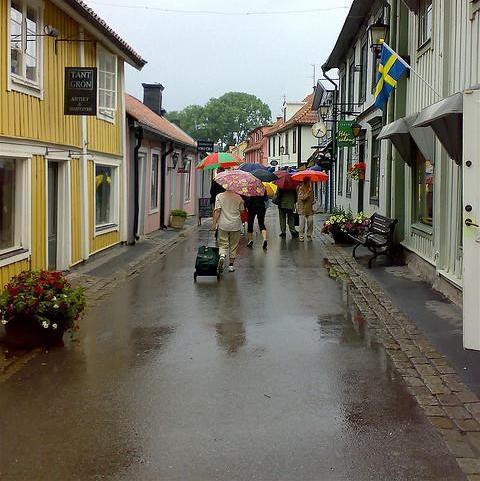}
		\includegraphics[width=0.15\linewidth]{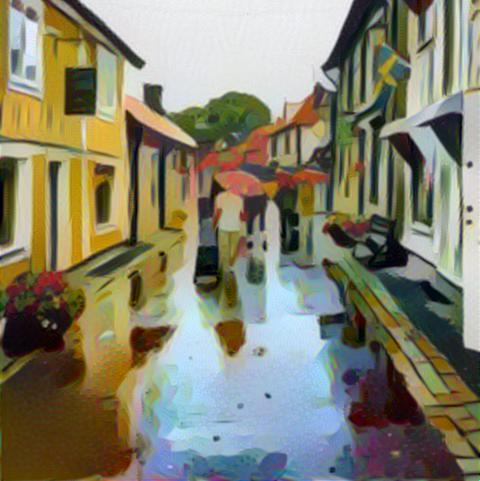}
		\includegraphics[width=0.15\linewidth]{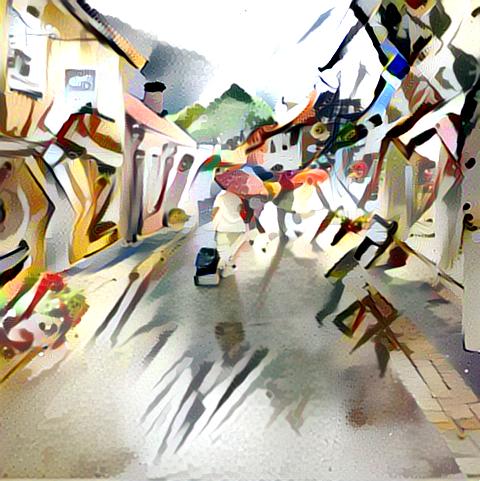}
		\includegraphics[width=0.15\linewidth]{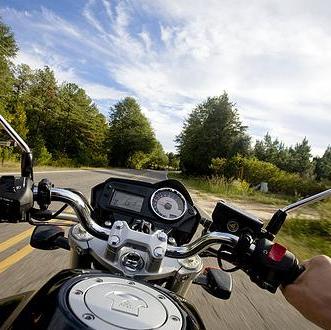}
		\includegraphics[width=0.15\linewidth]{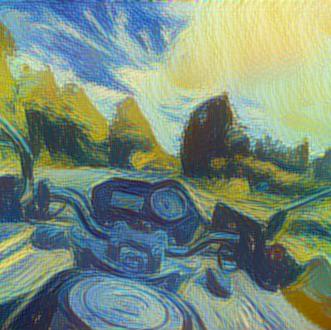}
		\includegraphics[width=0.15\linewidth]{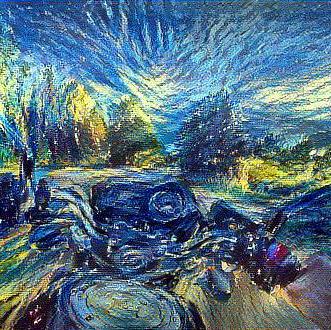}
	\end{subfigure}

	\begin{subfigure}[b]{0.5\linewidth}
		\hspace{10mm}
		\raisebox{7mm}{
			\begin{minipage}{30mm}
				\centering
				\textbf{Style} \\
				\textit{Composition X}, Wassily Kandinsky, 1939
			\end{minipage}
		}
		\includegraphics[height=17mm]{images/wassily_kandinsky.jpg}
	\end{subfigure}%
	\begin{subfigure}[b]{0.5\linewidth}
		\hspace{4mm}
		\raisebox{7mm}{
			\begin{minipage}{30mm}
				\centering
				\textbf{Style} \\
				\textit{Mountainprism}, Renee Nemerov, 2007
			\end{minipage}
		}
		\includegraphics[height=17mm]{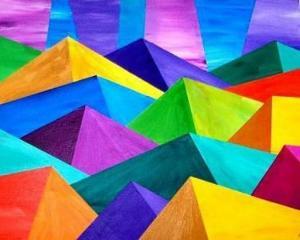}
	\end{subfigure}\\
	\begin{subfigure}[b]{1\linewidth}
		\centering
		\includegraphics[width=0.15\linewidth]{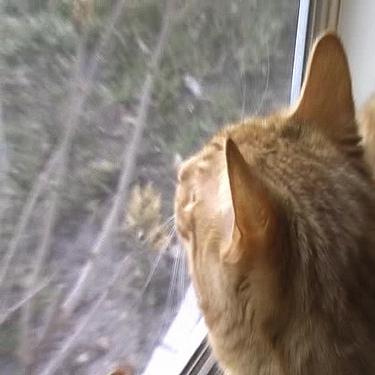}
		\includegraphics[width=0.15\linewidth]{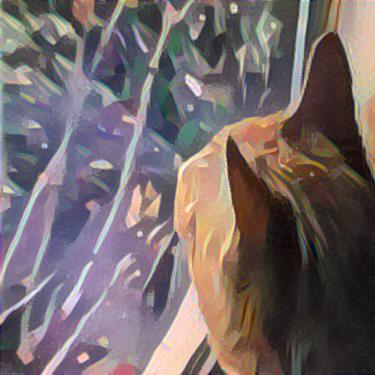}
		\includegraphics[width=0.15\linewidth]{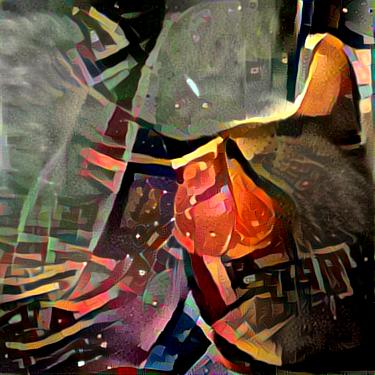}
		\includegraphics[width=0.15\linewidth]{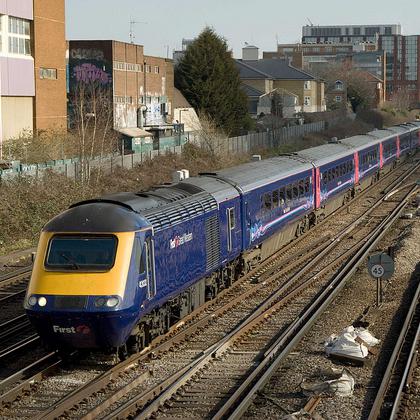}
		\includegraphics[width=0.15\linewidth]{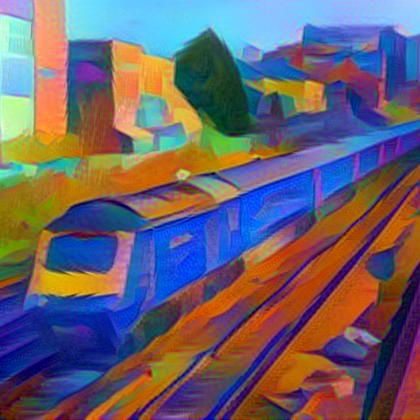}
		\includegraphics[width=0.15\linewidth]{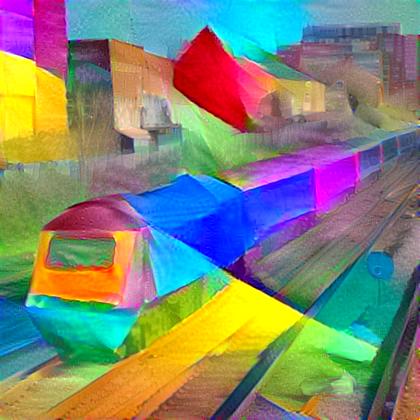}
	\end{subfigure} \\
	\vspace{1px}
	\begin{subfigure}[b]{1\linewidth}
		\centering
		\includegraphics[width=0.15\linewidth]{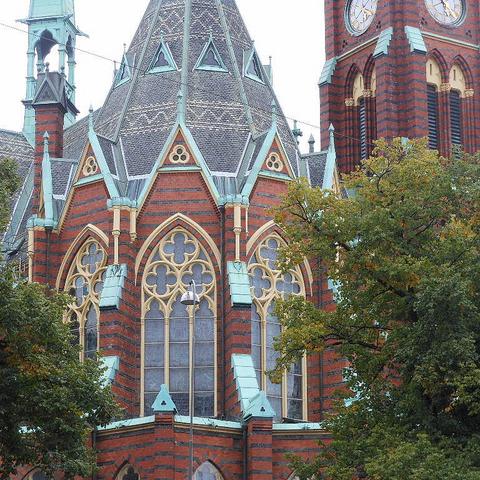}
		\includegraphics[width=0.15\linewidth]{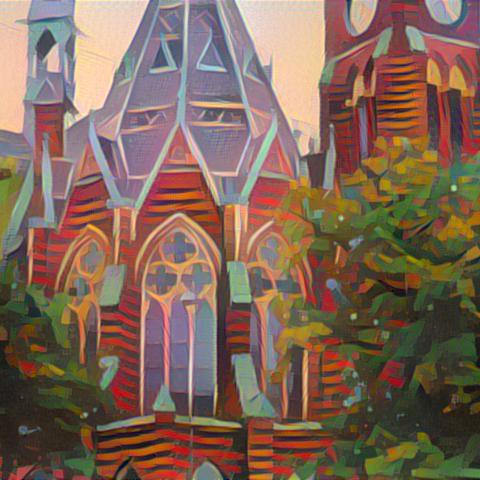}
		\includegraphics[width=0.15\linewidth]{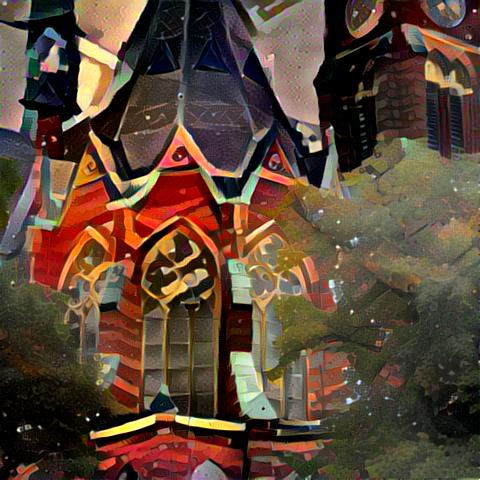}
		\includegraphics[width=0.15\linewidth]{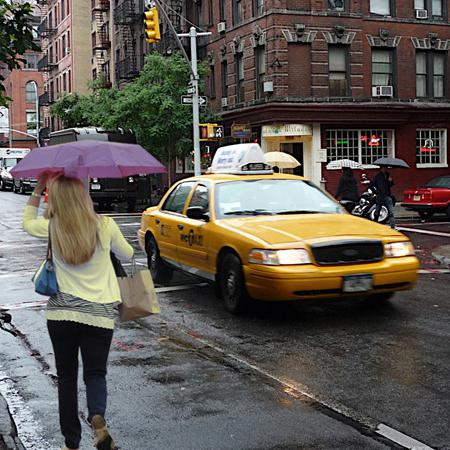}
		\includegraphics[width=0.15\linewidth]{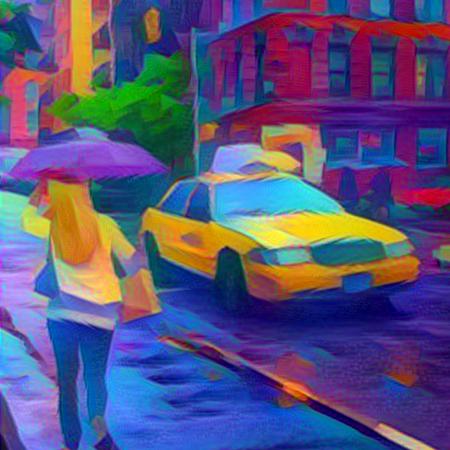}
		\includegraphics[width=0.15\linewidth]{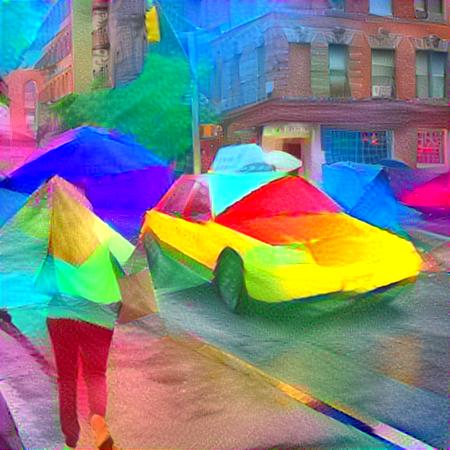}
	\end{subfigure}

	\begin{subfigure}[b]{0.5\linewidth}
		\hspace{10mm}
		\raisebox{7mm}{
			\begin{minipage}{30mm}
				\centering
				\textbf{Style} \\
				\textit{Butterfly Drawing}
			\end{minipage}
		}
		\includegraphics[height=17mm]{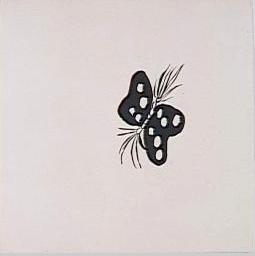}
	\end{subfigure}%
	%\begin{subfigure}[b]{0.5\linewidth}
	%\hspace{4mm}
	%\raisebox{7mm}{
	%\begin{minipage}{30mm}
	%      \centering
	%      \textbf{Style} \\
	%      \textit{Pencil Drawing}
	%    \end{minipage}
	%  }
	%\includegraphics[height=17mm]{images/pencilrose.jpg}
	%\end{subfigure}\\
	\begin{subfigure}[b]{0.5\linewidth}
		\hspace{4mm}
		\raisebox{7mm}{
			\begin{minipage}{30mm}
				\centering
				\textbf{Style} \\
				\textit{La Muse}, \\Pablo Picasso, \\1935
			\end{minipage}
		}
		\includegraphics[height=17mm]{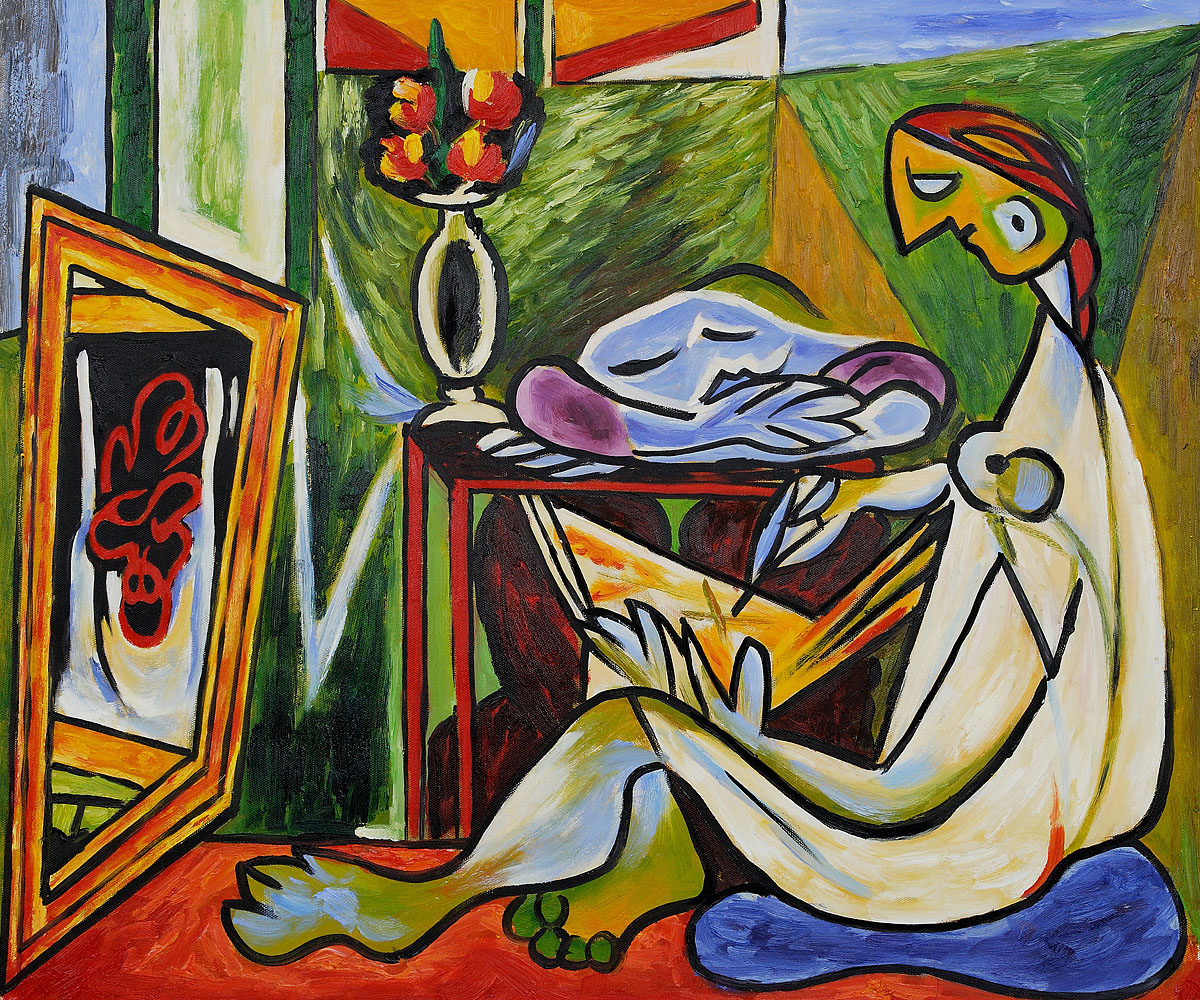}
	\end{subfigure}\\
	\begin{subfigure}[b]{1\linewidth}
		\centering
		\includegraphics[width=0.15\linewidth]{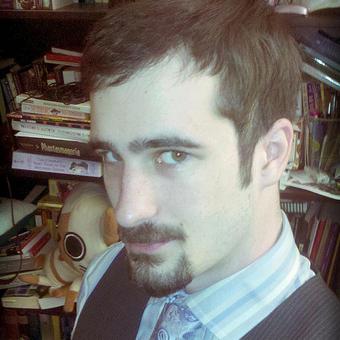}
		\includegraphics[width=0.15\linewidth]{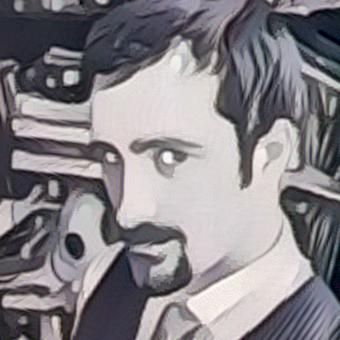}
		\includegraphics[width=0.15\linewidth]{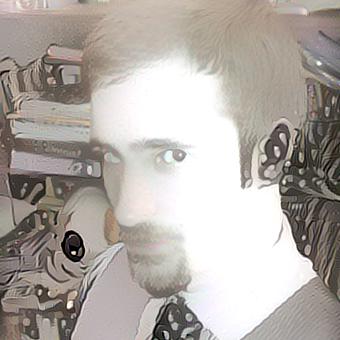}
		\includegraphics[width=0.15\linewidth]{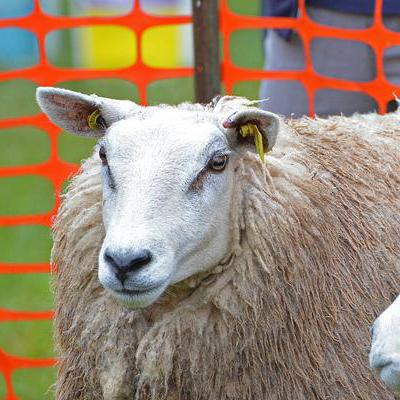}
		\includegraphics[width=0.15\linewidth]{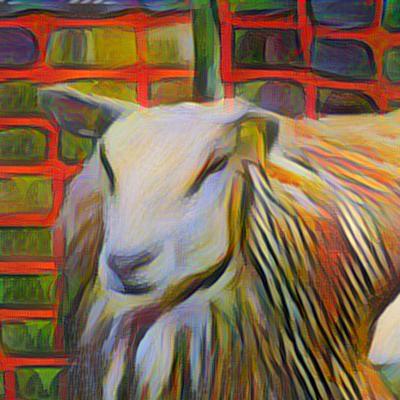}
		\includegraphics[width=0.15\linewidth]{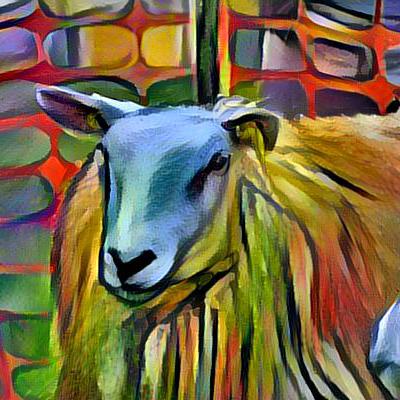}
	\end{subfigure} \\
	\vspace{1px}
	\begin{subfigure}[b]{1\linewidth}
		\centering
		\begin{subfigure}[b]{0.15\linewidth}
			\includegraphics[width=1\linewidth]{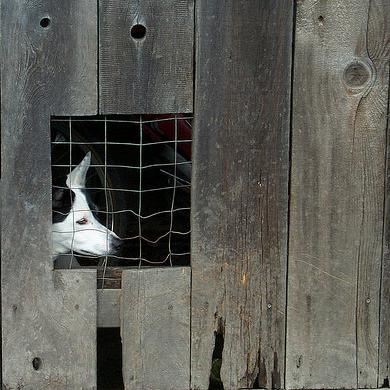}
			\caption*{\textbf{Content}}
		\end{subfigure}
		\begin{subfigure}[b]{0.15\linewidth}
			\includegraphics[width=1\linewidth]{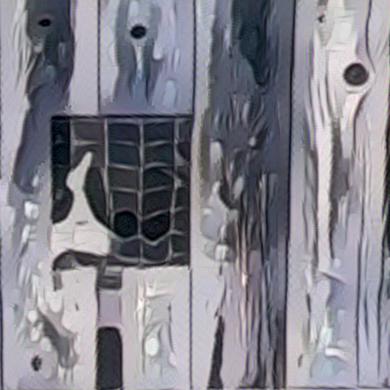}
			\caption*{Ours}
		\end{subfigure}
		\begin{subfigure}[b]{0.15\linewidth}
			\includegraphics[width=1\linewidth]{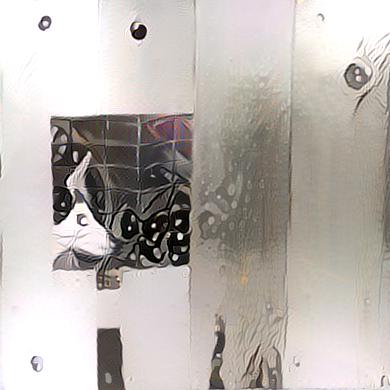}
			\caption*{Gatys \etal}
		\end{subfigure}
		%\begin{subfigure}[b]{0.15\linewidth}
		%\includegraphics[width=1\linewidth]{images/biker_orig.png}
		%\caption*{\textbf{Content}}
		%\end{subfigure}
		%\begin{subfigure}[b]{0.15\linewidth}
		%\includegraphics[width=1\linewidth]{images/biker_pencilrose.png}
		%\caption*{Ours}
		%\end{subfigure}
		%\begin{subfigure}[b]{0.15\linewidth}
		%\includegraphics[width=1\linewidth]{images/biker_pencilrose_gatys.jpg}
		%\caption*{Gatys \etal}
		%\end{subfigure}
		\begin{subfigure}[b]{0.15\linewidth}
			\includegraphics[width=1\linewidth]{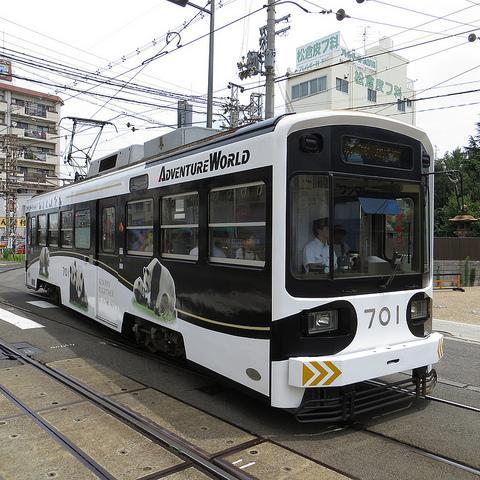}
			\caption*{\textbf{Content}}
		\end{subfigure}
		\begin{subfigure}[b]{0.15\linewidth}
			\includegraphics[width=1\linewidth]{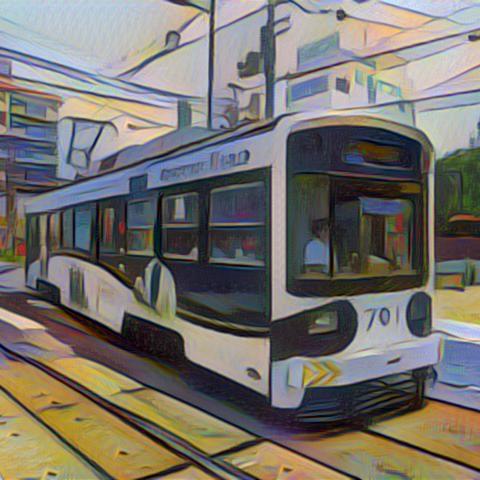}
			\caption*{Ours}
		\end{subfigure}
		\begin{subfigure}[b]{0.15\linewidth}
			\includegraphics[width=1\linewidth]{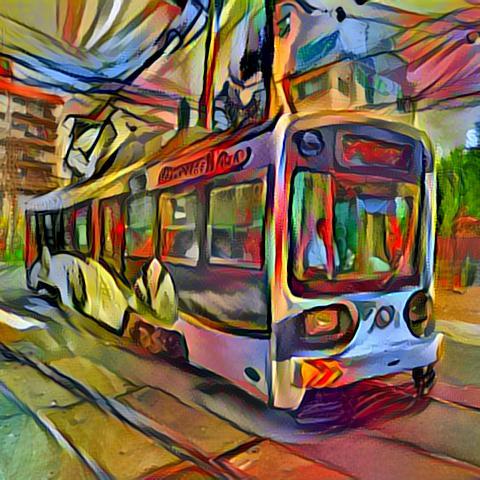}
			\caption*{Gatys \etal}
		\end{subfigure}
	\end{subfigure}
	
	\caption{Qualitative examples of our method compared with Gatys \etal's formulation of artistic style transfer. We show results using $3\times 3$ patches (most content-similar) and with the default configuration for Gatys \etal \cite{GatysEB15a}.}
	\label{fig:style-results}
\end{figure*}

\textbf{Target Layer.} The effects of style swapping in different layers of the VGG-19 network are shown in Figure~\ref{fig:swaplayers}. In this figure the RGB images are  computed by optimization as described in Section \ref{sec:method}. We see that while we can style swap directly in the RGB space, the result is nothing more than a recolor. As we choose a target layer that is deeper in the network, textures of the style image are more pronounced. We find that style swapping on the ``relu3\_1'' layer provides the most visually pleasing results, while staying structurally consistent with the content. We restrict our method to the ``relu3\_1'' layer in the following experiments and in the inverse network training. Qualitative results are shown in Figure~\ref{fig:style-results}.

\textbf{Consistency.} Our style swapping approach concatenates the content and style information into a single target feature vector, resulting in an easier optimization formulation compared to other approaches. As a result, we find that the optimization algorithm is able to reach the optimum of our formulation in less iterations than existing formulations while consistently reaching the same optimum. Figures~\ref{fig:consistency}~and~\ref{fig:quantconsistency} show the difference in optimization between our formulation and existing works under random initializations. Here we see that random initializations have almost no effect on the stylized result, indicating that we have far fewer local optima than other style transfer objectives.

\textbf{Straightforward Adaptation to Video.} This consistency property is advantageous when stylizing videos frame by frame. Frames that are the same will result in the same stylized result, while consecutive frames will be stylized in similar ways. As a result, our method is able to adapt to video without any explicit gluing procedure like optical flow \cite{Ruder2016}. We place stylized videos in the code repository.

\begin{figure}
	\begin{subfigure}[b]{0.33\linewidth}
		\centering
		\includegraphics[width=\linewidth]{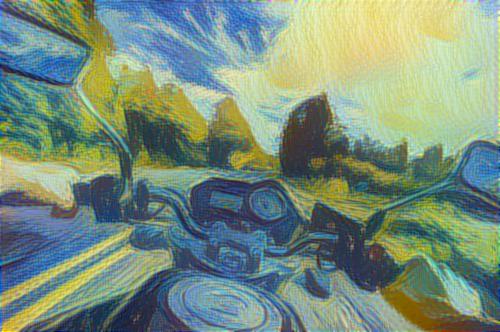}
	\end{subfigure}%
	\begin{subfigure}[b]{0.33\linewidth}
		\centering
		\includegraphics[width=\linewidth]{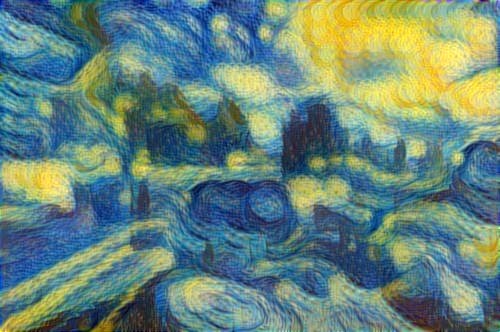}
	\end{subfigure}%
	\begin{subfigure}[b]{0.33\linewidth}
		\centering
		\includegraphics[width=\linewidth]{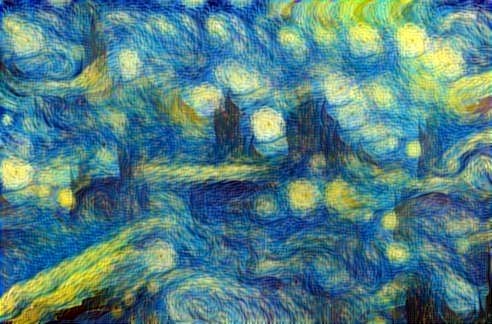}
	\end{subfigure}\\
	\begin{subfigure}[b]{0.33\linewidth}
		\centering
		\includegraphics[width=\linewidth]{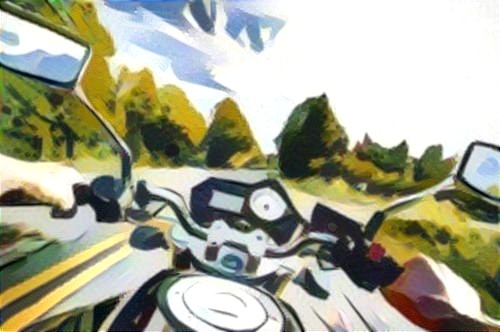}
		\caption*{patch size $3\times 3$}
	\end{subfigure}%
	\begin{subfigure}[b]{0.33\linewidth}
		\centering
		\includegraphics[width=\linewidth]{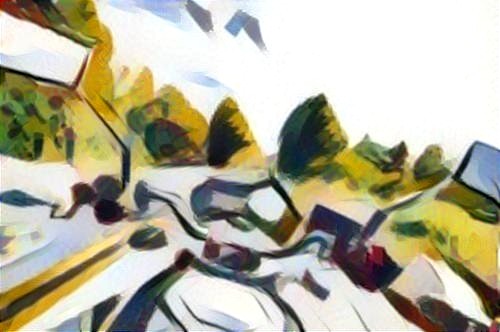}
		\caption*{patch size $7\times 7$}
	\end{subfigure}%
	\begin{subfigure}[b]{0.33\linewidth}
		\centering
		\includegraphics[width=\linewidth]{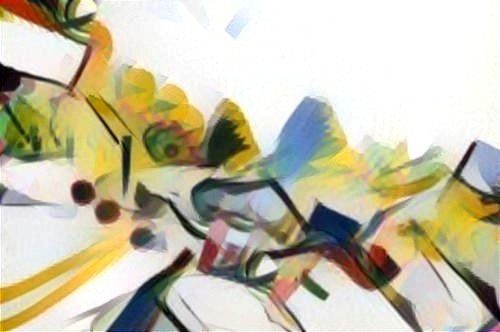}
		\caption*{patch size $12\times 12$}
	\end{subfigure}
	\caption{We can tradeoff between content structure and style texture by tuning the patch size. The style images, \textit{Starry Night} (top) and \textit{Small Worlds I} (bottom), are shown in Figure~\ref{fig:style-results}.}
	\label{fig:tuning}
\end{figure}

\textbf{Simple Intuitive Tuning.} 
A natural way to tune the degree of stylization (compared to preserving the content) in the proposed approach is to modify the patch size. Figure~\ref{fig:tuning} qualitatively shows the relationship between patch size and the style-swapped result. As the patch size increases, more of the structure of the content image is lost and replaced by textures in the style image. 

\subsection{CNN Inversion}
Here we describe our training of an inverse network that computes an approximate inverse function of the pretrained VGG-19 network \cite{simonyan2014very}. More specifically, we invert the truncated network from the input layer up to layer ``relu3\_1''. The network architecture is placed in the appendix.

\textbf{Dataset.} We train using the Microsoft COCO (MSCOCO) dataset \cite{mscoco} and a dataset of paintings sourced from wikiart.org and hosted by Kaggle \cite{paintingsdataset}. Each dataset has roughly $80,000$ natural images and paintings, respectively. Since typically the content images are natural images and style images are paintings, we combine the two datasets so that the network can learn to recreate the structure and texture of both categories of images. Additionally, the explicit categorization of natural image and painting gives respective content and style candidates for the augmentation described in Section \ref{sec:augment}.

\textbf{Training.} We resize each image to $256 \times 256$ pixels (corresponding to activations of size $64\times 64$) and train for approximately 2 epochs over each dataset. Note that even though we restrict the size of our training images (and corresponding activations), the inverse network is fully convolutional and can be applied to arbitrary-sized activations after training. 

We construct each minibatch by taking 2 activation samples from natural images and 2 samples from paintings. We augment the minibatch with 4 style-swapped activations using all pairs of natural images and paintings in the minibatch. We calculate subgradients using backpropagation on~\eqref{eq:trainInvNet} with the total variance regularization coefficient $\lambda = 10^{-6}$ (the method is not particularly sensitive to this choice), and we update parameters of the network using the Adam optimizer \cite{kingma2014adam} with a fixed learning rate of $10^{-3}$. 

\begin{figure}
	\centering
	\includegraphics[width=0.9\linewidth,trim={120px 290px 130px 290px},clip]{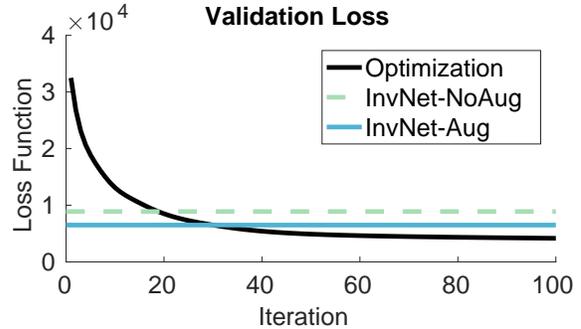}
	\caption{We compare the average loss \eqref{eq:trainInvNet} achieved by optimization and our inverse networks on 2000 variable-sized validation images and 6 variable-sized style images, using patch sizes of $3\times3$. Style images that appear in the paintings dataset were removed during training. }
	\label{fig:declosses}
\end{figure}

\textbf{Result.} Figure~\ref{fig:declosses} shows quantitative approximation results using 2000 full-sized validation images from MSCOCO and 6 full-sized style images. Though only trained on images of size $256 \times 256$, we achieve reasonable results for arbitrary full-sized images. We additionally compare against an inverse network that has the same architecture but was not trained with the augmentation. As expected, the network that never sees style-swapped activations during training performs worse than the network with the augmented training set.

\subsection{Computation Time}

\begin{table}
	\centering
	\ra{1.3}
	\begin{tabular}{@{}lrrr@{}}  
		\toprule
		Method & N. Iters. & Time/Iter. (s) & Total (s)\\
		\midrule
		Gatys \etal \cite{GatysEB15a} & 500 & 0.1004 & 50.20 \\
		Li and Wand \cite{Li2016}     & 200 & 0.6293 & 125.86\\
		Style Swap (Optim)            & 100 & 0.0466 & 4.66 \\
		Style Swap (InvNet)           & 1 & 1.2483 & 1.25 \\
		%Johnson \etal \cite{Johnson2016} & * & 0.0256 $\pm$ 0.0003 \\
		\bottomrule
	\end{tabular}
	\caption{Mean computation times of style transfer methods that can handle arbitary style images. Times are taken for images of resolution $300\times500$ on a GeForce GTX 980 Ti. Note that the number of iterations for optimization-based approaches should only be viewed as a very rough estimate.}
	\label{tab:time}
\end{table}

\begin{figure}[h]
	\begin{subfigure}[b]{0.49\linewidth}
		\includegraphics[width=\linewidth, trim= 130px 285px 155px 290px, clip]{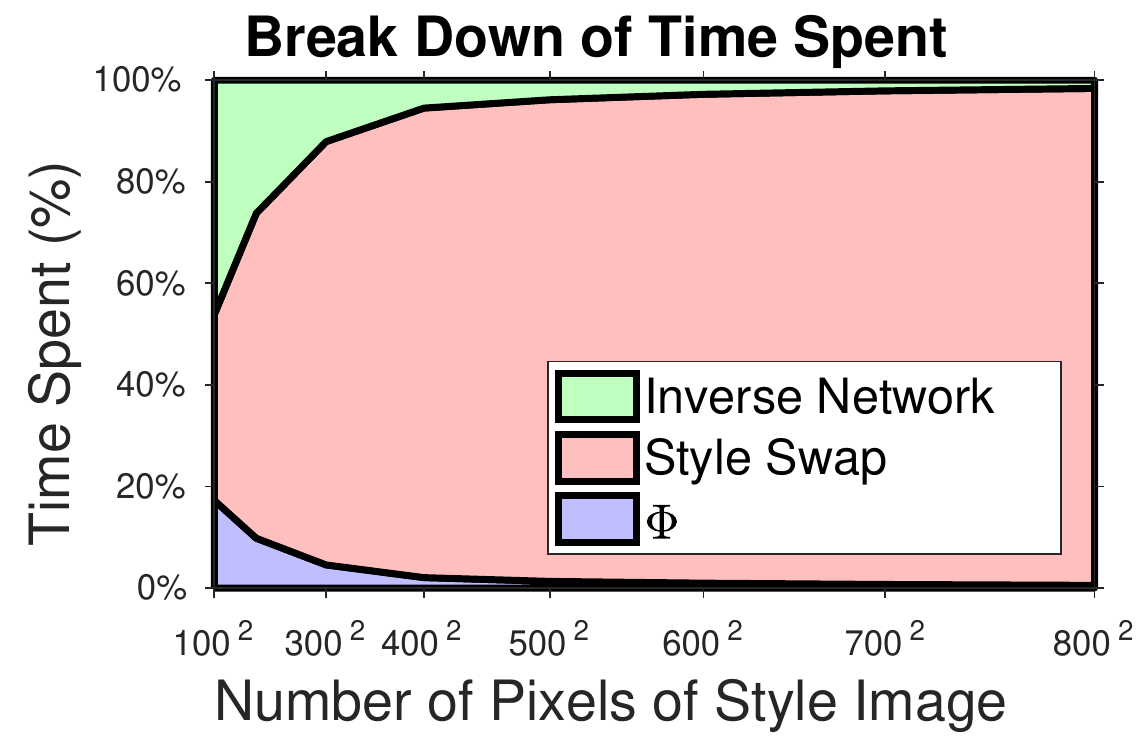}
		\vspace{-6mm}
		\caption{}
		\label{fig:styleperc}
		\vspace{1mm}
	\end{subfigure}%
	\begin{subfigure}[b]{0.49\linewidth}
		\includegraphics[width=\linewidth, trim= 130px 285px 155px 290px, clip]{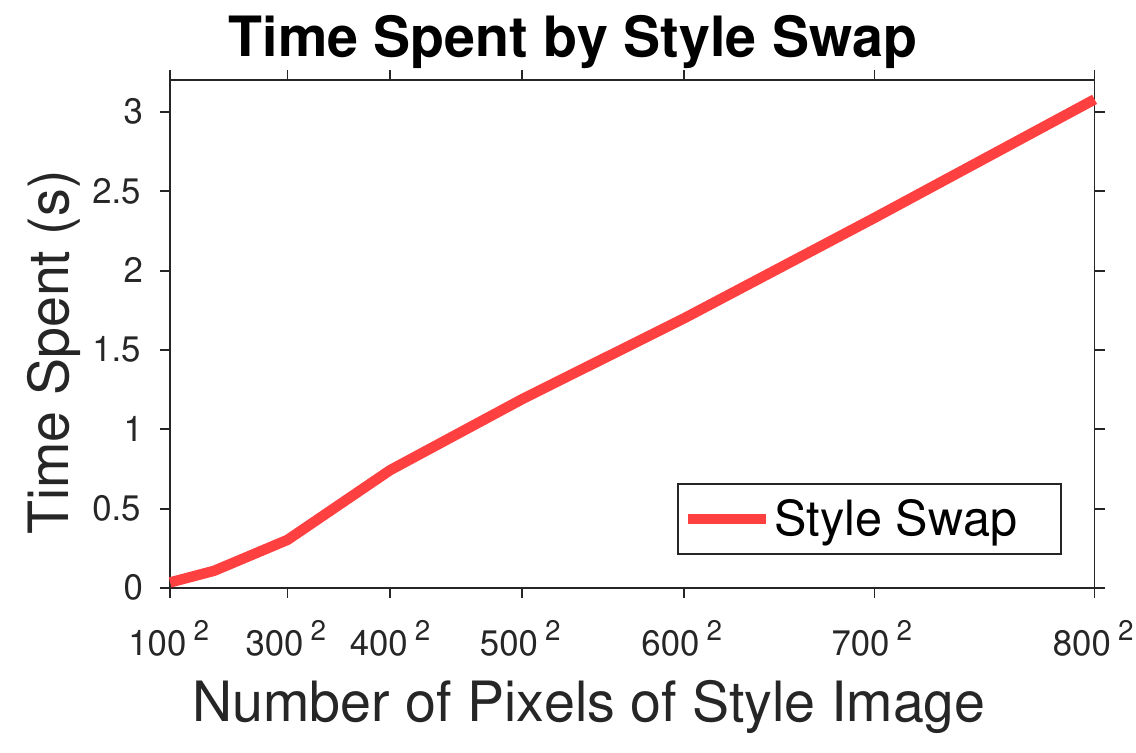}
		\vspace{-6mm}
		\caption{}
		\label{fig:stylelinear}
		\vspace{1mm}
	\end{subfigure}
	\begin{subfigure}[b]{0.49\linewidth}
		\includegraphics[width=\linewidth, trim= 130px 285px 155px 290px, clip]{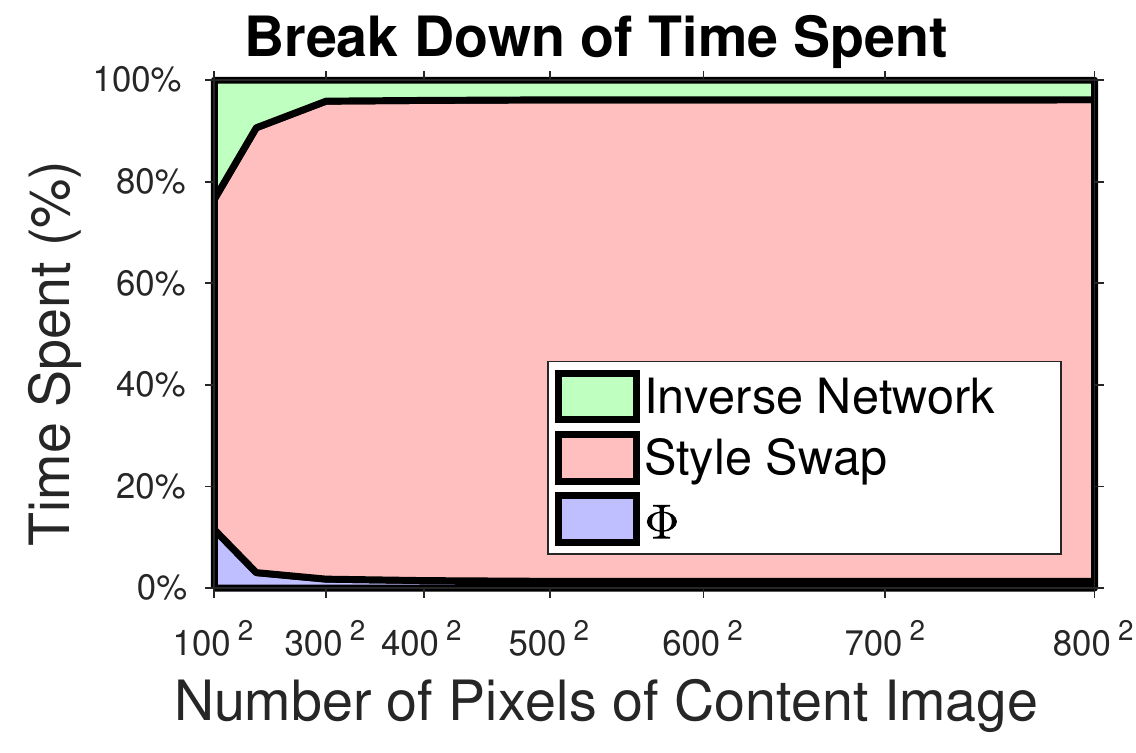}
		\vspace{-6mm}
		\caption{}
		\label{fig:contentperc}
		\vspace{1mm}
	\end{subfigure}%
	\begin{subfigure}[b]{0.49\linewidth}
		\includegraphics[width=\linewidth, trim= 130px 285px 155px 290px, clip]{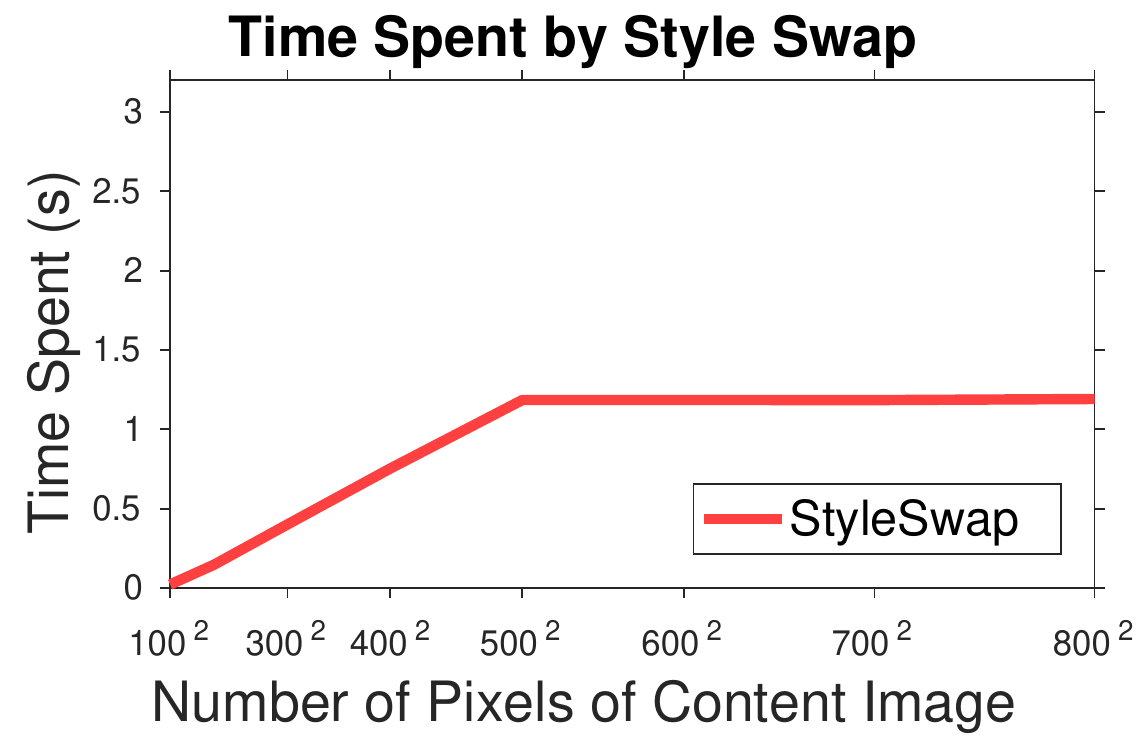}
		\vspace{-6mm}
		\caption{}
		\label{fig:contentlinear}
		\vspace{1mm}
	\end{subfigure}
	\caption{Compute times as (a,b) style image size increases and (c,d) as content image size increases. The non-variable image size is kept at $500\times500$. As shown in (a,c), most of the computation is spent in the style swap procedure. }
	\label{fig:computetimes}
\end{figure}

Computation times of existing style transfer methods are listed in Table~\ref{tab:time}. Compared to optimization-based methods, our optimization formula is easier to solve and requires less time per iteration, likely due to only using one layer of the pretrained VGG-19 network. Other methods use multiple layers and also deeper layers than we do.

%Compared to the feedforward method by Johnson \etal, we require more computation time. This is to due our method consisting of three parts (as described in Section~\ref{sec:ffstyle}) while Johnson \etal's procedure uses a single neural network that serves a similar role as our inverse network. 

%However, the biggest advantage of our method is the ability to stylize using any style image, whereas Johnson \etal's method requires training a new neural network for each style image. We've shown that our inverse method can generalize beyond its training set (Figure~\ref{fig:declosses}), while another useful application is the ability to change the size of the style image without retraining the inverse network. 

We show the percentage of computation time spent by different parts of our feedforward procedure in Figures~\ref{fig:styleperc}~and~\ref{fig:contentperc}. For any nontrivial image sizes, the style swap procedure requires much more time than the other neural networks. This is due to the style swap procedure containing two convolutional layers where the number of filters is the number of style patches. The number of patches increases linearly with the number of pixels of the image, with a constant that depends on the number of pooling layers and the stride at which the patches are extracted. Therefore, it is no surprise that style image size has the most effect on computation time (as shown in Figures~\ref{fig:styleperc}~and~\ref{fig:stylelinear}). 

Interestingly, it seems that the computation time stops increasing at some point even when the content image size increases (Figure~\ref{fig:contentlinear}), likely due to parallelism afforded by the implementation. This suggests that our procedure can handle large image sizes as long as the number of style patches is kept manageable. It may be desirable to perform clustering on the style patches to reduce the number of patches, or use alternative implementations such as fast approximate nearest neighbour search methods \cite{hajebi2011fast,muja2014scalable}.

%This suggests that it may be possible to further optimize for time efficiency when the style image is larger than the content image by using content patches as filters instead, as the normalized cross-correlation measure \eqref{eq:bestmatch} is symmetric.
\section{Discussion}\label{sec:discussion}

%We compare the compute times of different style transfer methods in Table~\ref{tab:time}. It can be readily seen that our optimization-based procedure is faster than existing ones. Our feedforward approach is slower than Johnson \etal \cite{Johnson2016} as their network 

We present a new CNN-based method of artistic style transfer that aims to be both fast and adaptable to arbitrary styles. Our method concatenates both content and style information into a single layer of the CNN, by swapping the textures of the content image with those of the style image. The simplistic nature of our method allows us to train an inverse network that can not only approximate the result in much less time, but can also generalize beyond its trained set of styles. Despite the one-layer restriction, our method can still produce visually pleasing images. Furthermore, the method has an intuitive tuning parameters (the patch size) and its consistency allows simple frame-by-frame application to videos.

While our feedforward method does not compete with the feedforward approach of Johnson \etal \cite{Johnson2016} in terms of speed, it should be noted that the biggest advantage of our method is the ability to stylize using new style images, whereas Johnson \etal's method requires training a new neural network for each style image. We've shown that our inverse method can generalize beyond its training set (Figure~\ref{fig:declosses}), while another useful application is the ability to change the size of the style image without retraining the inverse network. 

Some drawbacks of the proposed style swap procedure include lack of a global style measurement and lack of a similarity measure for neighboring patches, both in the spatial domain and the temporal domain. These simplifications sacrifice quality for efficiency, but can occassionally lead to a local flickering effect when applied to videos frame-by-frame. It may be desirable to look for ways to increase quality while keeping the efficient and versatile nature of our algorithm. 

%We proposed a style swap operation that uses a similarity measure to determine closest-matching patches. The closest-matching style patch for each content patch can be determined independently, and this allows parallelized implementation of the style swap using convolution layers. However, it may produce better spatial coherency to define closest-matching patches by also taking into account neighboring patches.

%We additionally showed that style swap is much more consistent than existing style transfer formulations. However, this consistency is not explicitly enforced by the style swap procedure, and frame-by-frame application can sometimes still produce high-frequency temporal changes. It may be desirable to also explicitly model temporal coherency in the style swap procedure, or use a similar gluing procedure to optical flow used in Ruder \etal \cite{Ruder2016}, to better adapt this method to videos. 

%We show qualitative comparisons with Gatys \etal in Figure~\ref{fig:style-results}, which is a state-of-the-art method for generating highly artistic stylizations. As our methods use different measures of ``artistic sense'', the stylized results are highly subjective. In particular, Gatys \etal's formulation results in images that are more abstract, while our method may find its use as a phone application for advanced photo filtering due to its consistency and speed. Regardless, it is possible to create more abstract artwork with our method by tuning the patch size as shown in Figure~\ref{fig:tuning}. 

{\small
\bibliographystyle{ieee}
\bibliography{ff-style-writeup}
}

\ifsupplementary

%%%%%%%%%% Merge with supplemental materials %%%%%%%%%%
\onecolumn
\begin{center}
	\textbf{\Large Appendix}
\end{center}
%%%%%%%%%% Merge with supplemental materials %%%%%%%%%%
%%%%%%%%%% Prefix a "S" to all equations, figures, tables and reset the counter %%%%%%%%%%
\setcounter{equation}{0}
\setcounter{figure}{0}
\setcounter{table}{0}
\setcounter{page}{1}
\makeatletter
\renewcommand{\theequation}{A\arabic{equation}}
\renewcommand{\thefigure}{A\arabic{figure}}
\renewcommand{\thetable}{A\arabic{table}}
\renewcommand{\thesection}{A\arabic{section}}
%%%%%%%%%% Prefix a "S" to all equations, figures, tables and reset the counter %%%%%%%%%%

\section*{Inverse Network Architecture}
The architecture of the truncated VGG-19 network used in the experiments is shown in Table~\ref{tab:vggnet}, and the inverse network architecture is shown in Table~\ref{tab:vgginv}. It is possible that better architectures achieve better results, as we did not try many different types of convolutional neural network architectures.

\begin{itemize}
	\renewcommand\labelitemi{--}
	\item Convolutional layers use filter sizes of $3\times 3$, padding of $1$, and stride of $1$.
	\item The rectified linear unit (ReLU) layer is an elementwise function $\text{ReLU}(x)=\max\{x,0\}$.
	\item The instance norm (InstanceNorm) layer standardizes each feature channel independently to have $0$ mean and a standard deviation of $1$. This layer has shown impressive performance in image generation networks \cite{DBLP:UlyanovVL16}.
	\item Maxpooling layers downsample by a factor of $2$ by using filter sizes of $2\times 2$ and stride of $2$.
	\item Nearest neighbor (NN) upsampling layers upsample by a factor of $2$ by using filter sizes of $2\times 2$ and stride of $2$.
\end{itemize}

\begin{table}[h]
	\centering
	\ra{1.3}
	\parbox{.40\linewidth}{
		\centering
		\begin{tabular}{@{} l c @{}}
			\toprule
			Layer Type & Activation Dimensions \\
			\midrule
			Input & $ H \times W \times 3$\\
			Conv-ReLU & $H \times W\times 64$ \\
			Conv-ReLU & $H \times W \times 64$ \\
			MaxPooling & $\nicefrac{1}{2}H \times \nicefrac{1}{2}W \times 64$ \\
			Conv-ReLU & $\nicefrac{1}{2}H \times \nicefrac{1}{2}W \times 128$ \\
			Conv-ReLU & $\nicefrac{1}{2}H \times \nicefrac{1}{2}W \times 128$ \\
			MaxPooling & $\nicefrac{1}{4}H \times \nicefrac{1}{4}W \times 128$ \\
			Conv-ReLU & $\nicefrac{1}{4}H \times \nicefrac{1}{4}W \times 256$ \\
			\bottomrule
		\end{tabular}
		\caption{Truncated VGG-19 network from the input layer to ``relu3\_1'' (last layer in the table).}
		\label{tab:vggnet}
	}
	\hspace{10mm}
	\parbox{.45\linewidth}{
		\centering
		\begin{tabular}{@{} l c @{}}
			\toprule
			Layer Type & Activation Dimensions \\
			\midrule
			Input & $\nicefrac{1}{4}H \times \nicefrac{1}{4}W \times 256$\\
			Conv-InstanceNorm-ReLU & $\nicefrac{1}{4}H \times \nicefrac{1}{4}W \times 128$ \\
			NN-Upsampling & $\nicefrac{1}{2}H \times \nicefrac{1}{2}W \times 128$ \\
			Conv-InstanceNorm-ReLU & $\nicefrac{1}{2}H \times \nicefrac{1}{2}W \times 128$ \\
			Conv-InstanceNorm-ReLU & $\nicefrac{1}{2}H \times \nicefrac{1}{2}W \times 64$ \\
			NN-Upsampling & $H \times W \times 64$ \\
			Conv-InstanceNorm-ReLU & $H \times W \times 64$ \\
			Conv & $H \times W \times 3$ \\
			\bottomrule
		\end{tabular}
		\caption{Inverse network architecture used for inverting activations from the truncated VGG-19 network.}
		\label{tab:vgginv}
	}
\end{table}

\fi

\end{document}